\pdfoutput=1
\documentclass[11pt]{article}
\usepackage{acl}
\usepackage{times}
\usepackage{hyperref}
\usepackage{latexsym}
\usepackage{amsthm}
\usepackage{amsfonts}
\usepackage{graphicx}
\usepackage{multirow}
\usepackage{pbox}
\usepackage{float}
\usepackage[export]{adjustbox}
\usepackage{graphicx}
 
\usepackage[font=small,skip=0pt]{caption}
\usepackage[T1]{fontenc}

\usepackage[utf8]{inputenc}

\usepackage{microtype}
\title{Towards Unification of Discourse Annotation Frameworks}

\author{Yingxue Fu \\
  School of Computer Science \\ University of St Andrews\\ KY16 9SX, UK \\
  \texttt{yf30@st-andrews.ac.uk} \\
 }

\begin{document}
\maketitle
\begin{abstract}
Discourse information is difficult to represent and annotate. Among the major frameworks for annotating discourse information, RST, PDTB and SDRT are widely discussed and used, each having its own theoretical foundation and focus. Corpora annotated under different frameworks vary considerably. To make better use of the existing discourse corpora and achieve the possible synergy of different frameworks, it is worthwhile to investigate the systematic relations between different frameworks and devise methods of unifying the frameworks. Although the issue of framework unification has been a topic of discussion for a long time, there is currently no comprehensive approach which considers unifying both discourse structure and discourse relations and evaluates the unified framework intrinsically and extrinsically. We plan to use automatic means for the unification task and evaluate the result with structural complexity and downstream tasks. We will also explore the application of the unified framework in multi-task learning and graphical models.
\end{abstract}

\section{Introduction}
A text is not a simple collection of isolated sentences. These sentences generally appear in a certain order and are connected with each other through logical or semantic means to form a coherent whole. In recent years, modelling beyond the sentence level is attracting more attention, and different natural language processing (NLP) tasks use discourse-aware models to obtain better performance, such as sentiment analysis~\citep{bhatia-etal-2015-better}, automatic essay scoring~\citep{nadeem-etal-2019-automated}, machine translation~\citep{sim-smith-2017-integrating}, text summarization~\citep{xu-etal-2020-discourse} and so on.

As discourse information typically involves the interaction of different levels of linguistic phenomena, including syntax, semantics, pragmatics and information structure, it is difficult to represent and annotate. Different discourse theories and discourse annotation frameworks have been proposed. Accordingly, discourse corpora annotated under different frameworks show considerable variation, and a corpus can be hardly used together with another corpus for natural language processing (NLP) tasks or discourse analysis in linguistics. Discourse parsing is a task of uncovering the underlying structure of text organization, and deep-learning based approaches are used in recent years. However, discourse annotation takes the whole document as the basic unit and is a laborious task. To boost the performance of neural models, we typically need a large amount of data. 

Due to the above issues, the unification of discourse annotation frameworks has been a topic of discussion for a long time. Researchers have proposed varied methods to unify discourse relations and debated over whether trees are a good representation of discourse~\citep{egg-redeker-2010-complex, lee2008departures, wolf-gibson-2005-representing}. However, existing research either focuses on mapping or unifying discourse relations of different frameworks~\citep{bunt2016iso, benamara-taboada-2015-mapping, sanders2018unifying, demberg2019compatible}, or on finding a common discourse structure~\citep{yi-etal-2021-unifying}, without giving sufficient attention to the issue of relation mapping. There is still no comprehensive approach that considers unifying both discourse structure and discourse relations. 

Another approach to tackling the task is to use multi-task learning so that information from a discourse corpus annotated under one framework can be used to solve a task in another framework, thus achieving synergy between different frameworks. However, existing studies adopting this method~\citep{liu2016implicit, braud-etal-2016-multi} do not show significant performance gain by incorporating a part of discourse information from a corpus annotated under a different framework. How to leverage discourse information from different frameworks remains a challenge.     

Discourse information may be used in down-stream tasks.~\citet{huang-kurohashi-2021-extractive} and~\citet{xu-etal-2020-discourse} use both coreference relations and discourse relations for text summarization with graph neural networks (GNNs). The ablation study by~\citet{huang-kurohashi-2021-extractive} shows that using coreference relations only brings little performance improvement but incorporating discourse relations achieves the highest performance gain. While different kinds of discourse information can be used, how to encode different types of discourse information to improve discourse-awareness of neural models is a topic that merits further investigation. 

The above challenges motivate our research on unifying different discourse annotation frameworks. We will focus on the following research questions:

\textbf{RQ1:} Which structure can be used to represent discourse in the unified framework? 

\textbf{RQ2:} What properties of different frameworks should be kept and what properties should be ignored in the unification?

\textbf{RQ3:} How can entity-based models and lexical-based models be incorporated into the unified framework? 

\textbf{RQ4:} How can the unified framework be evaluated?

The first three questions are closely related to each other. Automatic means will be used, although we do not preclude semi-automatic means, as exemplified by~\citet{yi-etal-2021-unifying}. We will start with the methods suggested by existing research and focus on the challenges of incorporating different kinds of discourse information in multi-task learning and graphical models. 

The unified framework can be used for the following purposes: 
\begin{enumerate}
  \item A corpus annotated under one framework can be used jointly with another corpus annotated under a different framework to augment data, for developing discourse parsing models or for discourse analysis.
  We can train a discourse parser on a corpus annotated under one framework and compare its performance with the case when it is trained on augmented data, similar to~\citet{yi-etal-2021-unifying}.  
  
  \item Each framework has its own theoretical foundation and focus. A unified framework may have the potential of combining the strengths of different frameworks. Experiments can be done with multi-task learning so that discourse parsing tasks of different frameworks can be solved jointly. We can also investigate how to enable GNNs to better capture different kinds of discourse information. 
  
  \item A unified framework may provide a common ground for exploring the relations of different frameworks and validating annotation consistency of a corpus. We can perform comparative corpus analysis and obtain new understanding of how information expressed in one framework is conveyed in another framework, thus validating corpus annotation consistency and finding some clues for solving problems in a framework with signals from another framework, similar to~\citet{polakova2017signalling} and~\citet{bourgonje-zolotarenko-2019-toward}.
\end{enumerate}

\section{Related Work}
\subsection{An Overview of Discourse Theories}
A number of discourse theories have been proposed. The theory by~\citet{grosz-sidner-1986-attention} is one of those earlier few whose linguistic claims about discourse are also computationally significant~\citep{mann1987rhetorical}. With this theory, it is believed that discourse structure is composed of three separated but interrelated components: linguistic structure, intentional structure and attentional structure. The linguistic structure focuses on cue phrases and discourse segmentation. The intentional structure mainly deals with why a discourse is performed (discourse purpose) and how a segment contributes to the overall discourse purpose (discourse segment purpose). The attentional structure is not related to the discourse participants, and it records the objects, properties and relations that are salient at each point in discourse. These three aspects capture discourse phenomena in a systematic way, and other discourse theories may be related to this theory in some way. For instance, the Centering Theory~\citep{grosz-etal-1995-centering} and the entity-grid model~\citep{barzilay-lapata-2008-modeling} focus on the attentional structure, and the Rhetorical Structure Theory (RST)~\citep{mann1988rhetorical} focuses on the intentional structure.

The theory proposed by~\citet{halliday2014cohesion} studies how various lexical means are used to achieve cohesion, these lexical means including reference, substitution, ellipsis, lexical cohesion and conjunction. Cohesion realized through the first four lexical means is in essence anaphoric dependency and conjunction is the only source of discourse relation under this theory~\citep{webber2006accounting}. 

The other discourse theories can be divided into two broad types: relation-based discourse theories and entity-based discourse theories~\citep{jurafsky2018speech}. The former studies how coherence is achieved with discourse relations and the latter focuses on local coherence achieved through shift of focus, which abstracts a text into a set of entity transition sequences~\citep{barzilay-lapata-2008-modeling}.

\begin{figure*}[h!]
\vspace{-6\baselineskip}
\noindent\begin{minipage}{\linewidth}
\centering
\includegraphics[
  width=0.9\textwidth,height=0.4\textheight, scale=1.2]{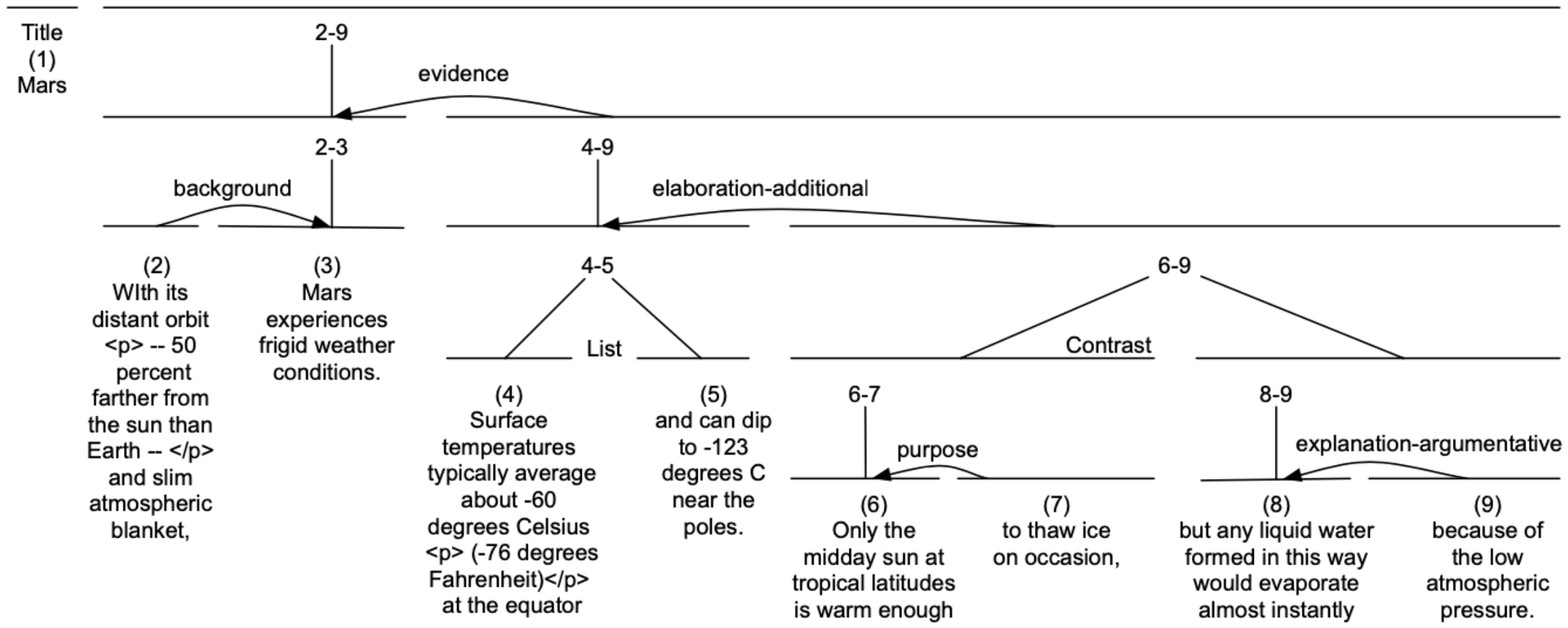}
 \vspace{-4\baselineskip}
  \caption{An RST discourse tree, originally from~\citet{marcu-2000-rhetorical}.}
  \label{rst-tree}
  
\end{minipage}
\end{figure*}

RST is one of the most influential relation-based discourse theories. The RST Discourse Treebank (RST-DT)~\citep{carlson-etal-2001-building} is annotated based on this theory. In the RST framework, discourse can be represented by a tree structure whose leaves are Elementary Discourse Units (EDUs), typically clauses, and whose non-terminals are adjacent spans linked by discourse relations. The discourse relations can be symmetric or asymmetric, the former being characterized by equally important spans connected in parallel, and the latter typically having a nucleus and a satellite, which are assigned based on their importance in conveying the intended effects. An RST tree is built recursively by connecting the adjacent discourse units, forming a hierarchical structure covering the whole text. An example of RST discourse trees can be seen in Figure~\ref{rst-tree}.

Another influential framework is the Penn Discourse Treebank (PDTB) framework, which is represented by the Penn Discourse Treebank~\citep{prasad-etal-2008-penn, prasad-etal-2018-discourse}. Unlike the RST framework, the PDTB framework does not aim at achieving complete annotation of the text but focuses on local discourse relations anchored by structural connectives or discourse adverbials. When there are no explicit connectives, the annotators will read adjacent sentences and decide if a connective can be inserted to express the relation. The annotation is not committed to a specific structure at the higher level. PDTB 3.0 adopts a three-layer sense hierarchy, including four general categories called classes at the highest level, the middle layer being more specific divisions, which are called types, and the lowest layer containing directionality of the arguments, called subtypes. An example of the PDTB-style annotation is shown as follows~\citep{ldcexample}:

\textit{The Soviet insisted that aircraft be brought into the talks,}(implicit=but)\{arg2-as-denier\}~\textbf{then argued for exempting some 4,000 Russian planes because they are solely defensive.} 

The first argument is shown in italics and the second argument is shown in bold font for distinction. As the discourse relation is implicit, the annotator adds a connective that is considered to be suitable for the context.

\begin{figure}
\begin{center}
\hbox{\hspace{-7.5 em}
  \includegraphics[
  width=0.9\textwidth,height=0.4\textheight, scale=1.2]
  {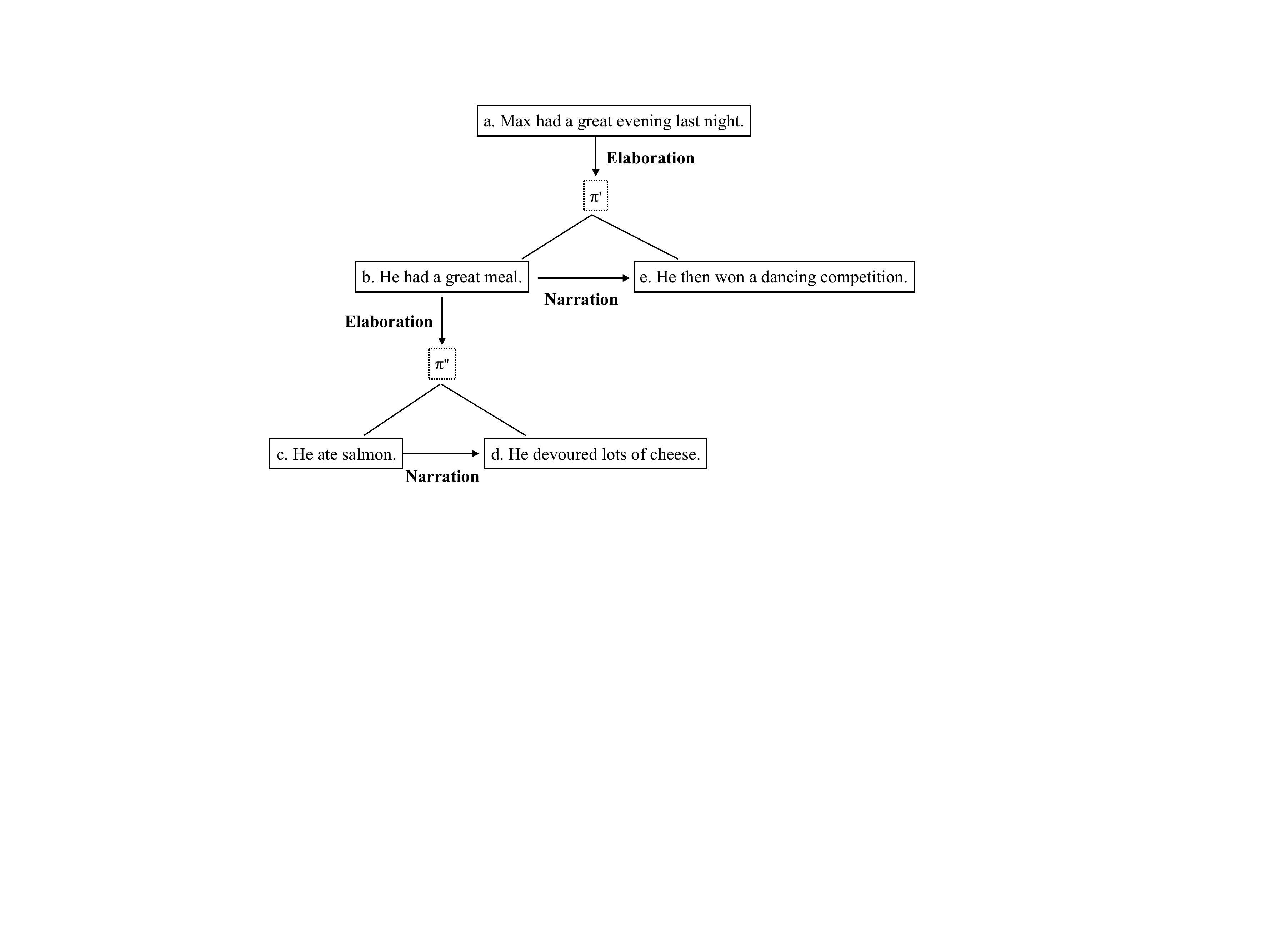}}
    \vspace{-9\baselineskip}
  \caption{SDRT representation of the text~\textit{a. Max had a great evening last night. b. He had a great meal. c. He ate salmon. d. He devoured lots of cheese. e. He then won a dancing competition.} The example is taken from~\citet{asher2003logics}.}
  \label{sdrt-graph}
\end{center}
\end{figure}

The Segmented Discourse Representation Theory (SDRT)~\citep{asher2003logics} is based on the Discourse Representation Theory~\citep{Kamp1993-KAMFDT}, with discourse relations added, and discourse structure is represented with directed acyclic graphs (DAGs). Elementary discourse units may be combined recursively to form a complex discourse unit (CDU), which can be linked with another EDU or CDU~\citep{asher2017annodis}. The set of discourse relations developed in this framework overlap partly with those in the RST framework but some are motivated from pragmatic and semantic considerations. In~\citet{asher2003logics}, a precise dynamic semantic interpretation of the rhetorical relations is defined. 
An example of discourse representation in the SDRT framework is shown in Figure~\ref{sdrt-graph}, which illustrates that the SDRT framework provides full annotation, similar to the RST framework, and it assumes a hierarchical structure of text organization. The vertical arrow-headed lines represent subordinate relations, and the horizontal lines represent coordinate relations. The textual units in solid-line boxes are EDUs and $\pi$\textquotesingle\ and $\pi$\textquotesingle\textquotesingle\ represent CDUs. The relations are shown in bold.

\subsection{Research on Relations between Different Frameworks}
The correlation between different frameworks has been a topic of interest for a long time. Some studies explore how different frameworks are related, either in discourse structures or in relation sets. Some studies take a step further and try to map the relation sets of different frameworks. 

\subsubsection{Comparison/unification of discourse structures of different frameworks}
~\citet{stede-etal-2016-parallel} investigate the relations between RST, SDRT and argumentation structure. For the purpose of comparing the three layers of annotation, the EDU segmentation in RST and SDRT is harmonized, and an ``argumentatively empty'' JOIN relation is introduced to address the issue that the basic unit of the argumentation structure is coarser than the other two layers. The annotations are converted to a common dependency graph format for calculating correlations. To transform RST trees to the dependency structure, the method introduced by~\citet{li-etal-2014-text} is used. The RST trees are binarized and the left-most EDU is treated as the head. In the transformation of the SDRT graphs to the dependency structure, the CDUs are simplified by a \textit{head replacement strategy}. The authors compare the dependency graphs in terms of common edges and common connected components. The relations of the argumentation structure are compared with those of RST and SDRT, respectively, through a co-occurrence matrix. Their research shows the systematic relations between the argumentation structure and the two discourse annotation frameworks. The purpose is to investigate if discourse parsing can contribute to automatic argumentation analysis. The authors exclude the PDTB framework because it does not provide full discourse annotation. 

~\citet{yi-etal-2021-unifying} try to unify two Chinese discourse corpora annotated under the PDTB framework and the RST framework, respectively, with a corpus annotated under the dependency framework. They use semi-automatic means to transform the corpora to the discourse dependency structure which is presented in~\citet{li-etal-2014-text}. Their work shows that the major difficulty is the transformation from the PDTB framework to the discourse dependency structure, which requires re-segmenting texts and complementing some relations to construct complete dependency trees. They use the same method as~\citet{stede-etal-2016-parallel} to transform the RST trees to the dependency structure. Details about relation mapping across the frameworks are not given. 

\subsubsection{Comparison/unification of discourse relations of different frameworks}

The methods of mapping discourse relations of different frameworks presented by~\citet{scheffler2016mapping}, ~\citet{demberg2019compatible} and~\citet{bourgonje-zolotarenko-2019-toward} are empirically grounded. The main approach is to make use of the same texts annotated under different frameworks.

~\citet{scheffler2016mapping} focus on mapping between explicit PDTB discourse connectives and RST rhetorical relations. The Potsdam Commentary Corpus~\citep{stede-neumann-2014-potsdam}, which contains annotations under both frameworks, is used. It is found that the majority of the PDTB connectives in the corpus match exactly one RST relation and mismatches are caused by different segment definitions and focuses, i.e., PDTB focuses on local/lexicalized relations and RST focuses on global structural relations.

As the Potsdam Commentary Corpus only contains explicit relations under the PDTB framework,~\citet{bourgonje-zolotarenko-2019-toward} try to induce implicit relations from the corresponding RST annotation. Since RST trees are hierarchical and the PDTB annotation is shallow, RST relations that connect complex spans are discarded. Moreover, because the arguments of explicit and implicit relations under the PDTB framework are determined based on different criteria, only RST relations that are signalled explicitly are considered in the experiment. It is shown that differences in segmentation and partially overlapping relations pose challenges for the task. 

~\citet{demberg2019compatible} propose a method of mapping RST and PDTB relations. Since the number of PDTB relations is much smaller than that of RST relations for the same text, the PDTB relations are used as the starting point for the mapping. They aim for mapping as many relations as possible while making sure that the relations connect the same segments. Six cases are identified: direct mapping, which is the easiest case; when PDTB arguments are non-adjacent, the Strong Compositionality hypothesis~\citep{marcu2000theory} (i.e., if a relation holds between two textual spans, that relation also holds between the most important units of the constituent spans) is used to check if there is a match when the complex span of an RST relation is traced along the nucleus path to its nucleus EDU; in the case of multi-nuclear relations, it is checked if a PDTB argument can be traced to the nucleus of the RST relation along the nucleus path; the mismatch caused by different segmentation granularity is considered innately unalignable and discarded; centrally embedded EDUs in RST-DT are treated as a whole and compared with an argument of the PDTB relation; and the PDTB E\textsc{nt}R\textsc{el} relation is included to test its correlation with some RST relations that tend to be associated with cohesion.  

Other studies are more theoretical.~\citet{hovy-1990-parsimonious} is the first to attempt to unify discourse relations proposed by researchers from different areas and suggests adopting a hierarchy of relations, with the top level being more general (from the functional perspective: ideational, interpersonal and textual) and putting no restrictions on adding fine-grained relations, as long as they can be subsumed under existing taxonomy. The number of researchers who propose a specific relation is taken as a vote of confidence of the relation in the taxonomy. The study serves as a starting point for research in this direction. There are a few other proposals for unifying discourse relations of different frameworks to facilitate cross-framework discourse analysis, including: introducing a hierarchy of discourse relations, similar to~\citet{hovy-1990-parsimonious}, where the top level is general and fixed, and the lowest level is more specific and allows variations based on genre and language~\citep{benamara-taboada-2015-mapping}, finding some dimensions based on cognitive evidence where the relations can be compared with each other and re-grouped~\citep{sanders2018unifying}, and formulating a set of core relations that are shared by existing frameworks but are open and extensible in use, with the outcome being ISO-DR-Core~\citep{bunt2016iso}. When the PDTB sense hierarchy is mapped to the ISO-DR-Core, it is found that the directionality of relations cannot be captured by the existing ISO-DR-Core relations and it remains a question whether to extend the ISO-DR-Core relations or to redefine the PDTB relations so that the directionality of arguments can be captured~\citep{prasad-etal-2018-discourse}. 

\section{Research Plan}

RST-DT is annotated on texts from the Penn Treebank~\citep{marcus-etal-1993-building} that have also been annotated in PDTB. The texts are formally written Wall Street Journal articles. The English corpora annotated under the SDRT framework, i.e., the STAC corpus~\citep{asher-etal-2016-discourse} and the Molweni corpus~\citep{li-etal-2020-molweni}, are created for analyzing multi-party dialogues, making it difficult to be used together with the other two corpora. Therefore, in addition to RST-DT and PDTB 3.0, we will use the ANNODIS corpus~\citep{pery-woodley-etal-2009-annodis}, which consists of formally written French texts. We will first translate the texts into English with an MT system and then manually check the translated texts to reduce errors. 

In the following, the research questions and the approach in our plan will be discussed. These questions are closely related to each other and the research on one question is likely to influence how the other questions should be addressed. They are presented separately just for easier description. 

\textbf{RQ1:} Which structure can be used to represent discourse in the unified framework? 

Although there is a lack of consensus on how to represent discourse structure, in a number of studies, the dependency structure is taken as a common structure that the other structures can be converted to~\citep{muller-etal-2012-constrained, hirao-etal-2013-single, venant-etal-2013-expressivity, li-etal-2014-text, yoshida-etal-2014-dependency, stede-etal-2016-parallel, morey-etal-2018-dependency, yi-etal-2021-unifying}. 
This choice is mainly inspired by research in the field of syntax, where the dependency grammar is better studied and its computational and representational properties are well-understood\footnote{In communication with Bonnie Webber, January, 2022.}. 
The research by~\citet{venant-etal-2013-expressivity} provides a common language for comparing discourse structures of different formalisms, which is used in the transformation procedure presented by~\citet{stede-etal-2016-parallel}. Another possibility is the constrained directed acyclic graph introduced by~\citet{danlos-2004-discourse}. While~\citet{venant-etal-2013-expressivity} focus on the expressivity of different structures, the constrained DAG is motivated from the perspective of strong generative capacity~\citep{danlos2008strong}. Although neither of the studies deals with the PDTB framework, since they are both semantically driven, we believe it is possible to deal with the PDTB framework using either of the two structures. We will start with the investigation of the two structures.

Another issue is how to maintain one-to-one correspondence during the transformation of the original structure and the unified structure back and forth. As indicated by~\citet{stede-etal-2016-parallel}, the transformation from the RST or SDRT structures into dependency structures always produces the same structure, but going back to the initial RST or SDRT structures is ambiguous.~\citet{morey-etal-2018-dependency} introduces head-ordered dependency trees in syntactic parsing~\citep{fernandez-gonzalez-martins-2015-parsing}
to reduce the ambiguity. We may start with a similar method. 

As is clear from Section 2, using the dependency structure as a common ground for studying the relations between different frameworks is not new in existing literature, but comparing the RST, PDTB and SDRT frameworks with this method has not yet been done. This approach will be our starting point, and the suitability of the dependency structure in representing discourse will be investigated empirically. The SciDTB corpus~\citep{yang-li-2018-scidtb}, which is annotated under the dependency framework, will be used for this purpose.

\textbf{RQ2:\footnote{In communication with Bonnie Webber, January, 2022. We thank her for pointing out this aspect.}} What properties of different frameworks should be kept and what properties should be ignored in the unification?

We present a non-exhaustive list of properties, which we consider to have considerable influence on the unified discourse structure.  
\begin{enumerate}
    \item Nuclearity:~\citet{marcu1996building} uses the nuclearity principle as the foundation for a formal treatment of compositionality in RST, which means that two adjacent spans can be joined into a larger span by a rhetorical relation if and only if the relation holds between the most salient units of those spans. This assumption is criticized by~\citet{stede2008disentangling}. The remedy provided by~\citet{stede2008disentangling} is to separate different levels of discourse information, which is in line with the suggestions in~\citet{Knott00beyondelaboration:} and~\citet{ moore-pollack-1992-problem}. Our strategy is to keep this property in the initial stage of experimentation. The existing methods for transforming RST trees to dependency structure~\citep{hirao-etal-2013-single, li-etal-2014-text} rely heavily on the nuclearity principle and we will use these methods in the transformation and see what kinds of problems this procedure will cause, particularly with respect to the PDTB framework, which does not enforce a hierarchical structure for complete coverage of the text. 
   
    \item Sentence-boundedness: 
    The RST framework does not enforce well-formed discourse sub-trees for each sentence. However, it is found that 95\% of the discourse parse trees in RST-DT have well-formed sub-trees at the sentence level~\citep{soricut-marcu-2003-sentence}. For the PDTB framework, there is no restriction on how far an argument can be from its corresponding connective: it can be in the same sentence as the connective, in the sentence immediately preceding that of the connective, or in some non-adjacent sentence~\citep{Prasad2006ThePD}. Moreover, the arguments are determined based on the \textit{Minimality Principle}, which means that clauses and/or sentences that are minimally required for the interpretation of the relation should be included in the argument, and other spans that are relevant but not necessary can be annotated as supplementary information, which is labeled depending on which argument it is supplementary to~\citep{prasad-etal-2008-penn}. The SDRT framework developed in~\citet{asher2003logics} does not specify the basic discourse unit, but in the annotation of the ANNODIS corpus, EDU segmentation follows similar principles as RST-DT. The formation of CDU and the attachment of relations are where SDRT differs significantly from RST. A segment can be attached to another segment from the same sentence, the same paragraph or a larger context, and by one or possibly more relations. A CDU can be of any size and can have segments that are far apart in the text, and relations may be annotated within the CDU\footnote{See section 3 of the ANNODIS annotation manual, available through \url{ http://w3.erss.univ-tlse2.fr/textes/publications/CarnetsGrammaire/carnGram21.pdf}}.
    
    The differences in the criteria on location and extent for basic discourse unit identification and relation labeling of the RST framework and the PDTB framework may be partly attributed to different annotation procedures. In RST, EDU segmentation is performed first and EDU linking and relation labelling are performed later. The balance between consistency and granularity is the major concern behind the strategy for EDU segmentation~\citep{carlson-etal-2001-building}. In contrast, in PDTB, the connectives are identified first, and their arguments are determined afterwards. Semantic relatedness is given greater weight and the location and extent of the arguments can be determined more flexibly. On the whole, neither SDRT nor PDTB shows any tendency of sentence-boundedness. We will investigate to what extent the tendency of sentence-boundedness complicates the unification and what the consequences are if entity-based models and lexical-based models are incorporated. 
    
    \item Multi-sense annotation: As shown above, SDRT and PDTB allow multi-sense annotation while RST only allows one relation to be labeled. The single-sense constraint actually gives rise to ambiguity because of the multi-faceted nature of local coherence~\citep{stede2008disentangling}. For the unification task, we assume that multi-sense annotation is useful. However, we agree with the view mentioned in~\citet{stede2008disentangling} that incrementally adding more relations as phenomena are being recognized is not a promising direction. There are two possible approaches: one is to separate different dimensions of discourse information~\citep{stede2008disentangling} and the other is to represent different kinds of discourse information simultaneously, similar to the approach adopted in~\citet{Knott00beyondelaboration:}. While multi-level annotation may reveal the interaction between discourse and other linguistic phenomena, it is less helpful for developing a discourse parser and requires more efforts in annotation. The second approach may be conducive to computationally cheaper discourse processing when proper constraints are introduced. 
\end{enumerate}

\textbf{RQ3:} How can entity-based models and lexical-based models be incorporated into the unified framework? 

The PDTB framework believes that lexical-based discourse relations are associated with anaphoric dependency, which is anchored by discourse adverbials~\citep{anaphora-discourse} and annotated as a type of explicit relations. As for entity-based relations, PDTB uses the E\textsc{nt}R\textsc{el} label to annotate this type of relations when neither explicit nor implicit relations can be identified and only entity-based coherence relations are present. In the RST framework, the ELABORATION relation is actually a relation between entities. However, it is encoded in the same way as the other relations between propositions, which bedevils the framework~\citep{Knott00beyondelaboration:}. Further empirical studies may be needed to identify how different frameworks represent these different kinds of discourse information. The main challenge is to use a relatively simple structure to represent different types of discourse information while keeping the complexity relatively low. 

\textbf{RQ4:} How can the unified framework be evaluated?

We will use intrinsic evaluation to assess the complexity of the discourse structure.

Extrinsic evaluation will be used to assess the effectiveness of the unified framework. The downstream tasks in the extrinsic evaluation include text summarization and document discrimination, which are two typical tasks for evaluating discourse models. The document discrimination task asks a score of coherence to be assigned to a document. The originally written document is considered to be the most coherent, and with more permutations, the document becomes less coherent. For comparison with previous studies, we will use the CNN and Dailymail dataset~\citep{cnndailymaildataset15} for the text summarization task, and use the method and dataset\footnote{\url{https://github.com/AiliAili/Coherence_Modelling}} in~\citet{shen-etal-2021-evaluating} to control the degree of coherence for the document discrimination task. 

Previous studies that use multi-task learning and GNNs to encode different types of discourse information will be re-investigated to test the effectiveness of the unified framework. 

As we may have to ignore some properties, we will examine what might be lost with the unified framework. 

\section{Conclusion}
We propose to unify the RST, PDTB and SDRT frameworks, which may enable discourse corpora annotated under different frameworks to be used jointly and achieve the potential synergy of different frameworks. The major challenges include determining which structure to use in the unified framework, choosing what properties to keep and what to ignore, and incorporating entity-based models and lexical-based models into the unified framework. We will start with existing research and try to find a computationally less expensive way for the task. Extensive experiments will be conducted to investigate how effective the unified framework is and how it can be used. An empirical evaluation of what might be lost through the unification will be performed. 

\section{Acknowledgements}
We thank Bonnie Webber for valuable feedback that greatly shaped the work. We are grateful to the anonymous reviewers for detailed and insightful comments that improved the work considerably, and Mark-Jan Nederhof for proof-reading the manuscript. The author is funded by University of St Andrews-China Scholarship Council joint scholarship (NO.202008300012).

\section{Ethical Considerations and Limitations}
The corpora are used in compliance with the licence requirements:

The ANNODIS corpus is available under Creative Commons By-NC-SA 3.0. 

RST-DT is distributed on Linguistic Data Consortium: 

Carlson, Lynn, Daniel Marcu, and Mary Ellen Okurowski. RST Discourse Treebank LDC2002T07. Web Download. Philadelphia: Linguistic Data Consortium, 2002.

PDTB 3.0 is also distributed on Linguistic Data Consortium: 

Prasad, Rashmi, et al. Penn Discourse Treebank Version 3.0 LDC2019T05. Web Download. Philadelphia: Linguistic Data Consortium, 2019.

\textbf{Bender Rule} English is the language studied in this work. 

\bibliography{anthology,custom}
\bibliographystyle{acl_natbib}

\end{document}